\pdfoutput=1

\documentclass[11pt]{article}

\usepackage{acl}
\usepackage{times}
\usepackage{latexsym}
\usepackage{graphicx}
\usepackage{amsmath}
\usepackage{booktabs}
\usepackage{afterpage}
\usepackage{subcaption}
\usepackage{makecell}
\usepackage{adjustbox}
\usepackage{algorithm}
\usepackage{algorithmic}
\usepackage{etoolbox}
\usepackage{enumitem}
\usepackage{multirow}
\usepackage{pifont}

\newcommand{\cmark}{\ding{51}} 
\newcommand{\xmark}{\ding{55}} 

\usepackage[T1]{fontenc}

\usepackage[utf8]{inputenc}

\usepackage{microtype}

\title{KIEval: A Knowledge-grounded Interactive Evaluation Framework for Large Language Models}

\author{
    \textbf{
        Zhuohao Yu\textsuperscript{{1}},
        Chang Gao\textsuperscript{{1}},
        Wenjin Yao\textsuperscript{{1}},
        Yidong Wang\textsuperscript{{1}},
    } \\
    \textbf{
        Wei Ye\textsuperscript{{1}}\thanks{\llap{}\:\:\:Corresponding author. },
        Jindong Wang\textsuperscript{{2}},
        Xing Xie\textsuperscript{{2}},
        Yue Zhang\textsuperscript{{3}},
        Shikun Zhang\textsuperscript{{1}}
    } \\
    \textsuperscript{1}Peking University
    \textsuperscript{2}Microsoft Research
    \textsuperscript{3}Westlake University \\
    \texttt{zyu@stu.pku.edu.cn, wye@pku.edu.cn}
}

\begin{document}
\maketitle

\begin{abstract}

	Automatic evaluation methods for large language models (LLMs) are hindered by data contamination, leading to inflated assessments of their effectiveness. Existing strategies, which aim to detect contaminated texts, focus on quantifying contamination status instead of accurately gauging model performance. In this paper, we introduce \textbf{KIEval}, a \underline{K}nowledge-grounded \underline{I}nteractive \underline{Eval}uation framework,  which incorporates an LLM-powered "interactor" role for the first time to accomplish a dynamic contamination-resilient evaluation. Starting with a question in a conventional LLM benchmark involving domain-specific knowledge, KIEval utilizes dynamically generated, multi-round, and knowledge-focused dialogues to determine whether a model's response is merely a recall of benchmark answers or demonstrates a deep comprehension to apply knowledge in more complex conversations. Extensive experiments on seven leading LLMs across five datasets validate KIEval's effectiveness and generalization. We also reveal that data contamination brings no contribution or even negative effect to models' real-world applicability and understanding, and existing contamination detection methods for LLMs can only identify contamination in pre-training but not during supervised fine-tuning.

\end{abstract}

\section{Introduction}
The landscape of artificial intelligence has been significantly reshaped by the emergence of Large Language Models (LLMs) as they have been pivotal in various natural language understanding and generation tasks~\citep{brown2020language, openai2023gpt4, bubeck2023sparks}. To better understand the capabilities and weaknesses of LLMs, their effective evaluation becomes increasingly essential~\citep{chang2023survey, guo2023evaluating}.

Automatic evaluation methods of LLMs generally fall into two categories: static dataset-based and LLM-based evaluation~\citep{chang2023survey}.
The former~\citep{clark2018arc, zellers2019hellaswag, hendrycks2020mmlu, huang2023ceval} requires evaluated LLMs to generate a short span of text containing choices or answers for pre-defined questions~\citep{eval-harness} to challenge their knowledge. The latter~\citep{chiang2023can}, also known as LLM-as-a-judge, typically depends on LLM evaluators to evaluate the model's outputs given predetermined questions or instructions~\citep{mtbench, lin2023llmeval, fu2023gptscore, pandalm}. Despite these promising efforts, current evaluation methodologies still broadly face the bottleneck of data contamination~\citep{schaeffer2023pretraining,wei2023skywork, oren2023proving,sainz2023nlp,daniele2023amplify-instruct}, where models trained on test splits of datasets can artificially inflate benchmark performance, overestimating their real-world efficacy and even potentially misleading scientific conclusions~\citep{zhou2023dont}.

Recently, two primary strategies have been employed to mitigate data contamination of LLMs. The first involves identifying whether specific texts or test samples exist in the training dataset by assessing loss values~\cite{wei2023skywork,shi2023detecting} or probing datasets like Common Craw~\cite{li2023open}. The limitation lies in its capacity to only measure contamination levels rather than actual model performance.
Meanwhile, this technique demands access to the model's internal structure or training datasets, rendering it ineffective for proprietary LLMs. The second strategy creates dynamic evaluation samples through heuristic methods, such as graph-based processes~\cite{zhu2023dyval}, yet this is confined to particular tasks (e.g., multi-step reasoning). Currently, there is a lack of a generalized evaluation protocol capable of assessing genuine performance amidst data contamination across diverse tasks and domains for both open and closed-source LLMs.

To this end, we propose \textbf{KIEval}, a \underline{K}nowledge-grounded \underline{I}nteractive \underline{Eval}uation framework, where a novel LLM-powered role, named "interactor," is introduced into the evaluation process for the first time. The term "knowledge-grounded" refers to our evaluation's starting point, which involves posing a question from an existing benchmark dataset that demands domain-specific knowledge. By "interactive," we mean the evaluation process delves deeper with structured and dynamic multi-round dialogues—tailored by the proposed interactor—to explore knowledge related to the initial question. These technical designs inherently provide our evaluation framework with two distinct merits.

\begin{itemize}[leftmargin=1em]
	\setlength\itemsep{0em}
	\item \textbf{Contamination-Resilient:}  KIEval marks a departure from conventional approaches that evaluate a model's capability in responding to static questions. Dynamic multi-round interactions allow us to distinguish whether a model's answer stems from a simple recall of benchmark answers or reflects a sound understanding to apply knowledge in problem-solving.

	\item \textbf{Generalized and Scalable:} Leveraging the capabilities of advanced LLMs as interactors renders our evaluation method universally applicable and eliminates the need for additional human efforts. Meanwhile, by reusing high-quality benchmark datasets as a foundation for domain knowledge, KIEval enables efficient scalability across diverse domains, tasks, and languages without significant resource expenditure.

\end{itemize}

We validate KIEval's alignment with humans and compare it against previous evaluation methods.
Our experiments show that KIEval achieves a high Pearson correlation coefficient of 0.81 with human scores, underscoring KIEval's proficiency in reflecting human preferences in our settings compared to previous evaluation methods. We also analyze KIEval's correlation with static dataset-based benchmarks, identifying that notable disparities in performance could signal data contamination.

Overall, our core contributions are three-fold:
\begin{itemize}[leftmargin=1em]
	\setlength\itemsep{0em}
	\item \emph{A novel dynamic evaluation protocol.} KIEval pioneeringly evaluates LLMs through dynamic multi-round interactions to mitigate data contamination. By seamlessly integrating with existing datasets as knowledge sources,  KIEval can cost-effectively assess knowledge memorization and generalization across domains and tasks.

	\item \emph{Extensive evaluation of popular LLMs.} We conduct thorough experiments and analysis with seven leading LLMs across five datasets with KIEval, assessing their generative abilities and domain knowledge, confirming the susceptibility of current evaluation methods (e.g., static dataset-based and LLM-based evaluations) to data contamination.

	\item \emph{New insights into data contamination.} Our investigation reveals the incompetence of data contamination in improving LLMs' genuine understanding and generalization, with current detection methods unable to identify contamination in the fine-tuning phase.

\end{itemize}

We release all necessary code and data for reproducing our method and the compared baselines.\footnote{We release all materials at~\url{https://github.com/zhuohaoyu/KIEval}.}

\section{Related Work}
\nocite{*}

\subsection{Evaluating LLMs}
\textbf{Human evaluation} approaches manually design experiments and tests~\citep{novikova2017we, bommasani2023holistic}. While it provides insights into human-model interaction, it faces challenges due to the subjectivity and inconsistency of human judgments \citep{chang2023survey}. Moreover, it is resource-intensive in terms of time and cost, limiting its feasibility for large-scale assessments \citep{karpinska2021perils}.

\begin{figure*}[t!]
	\centering
	\includegraphics[width=1.0\textwidth]{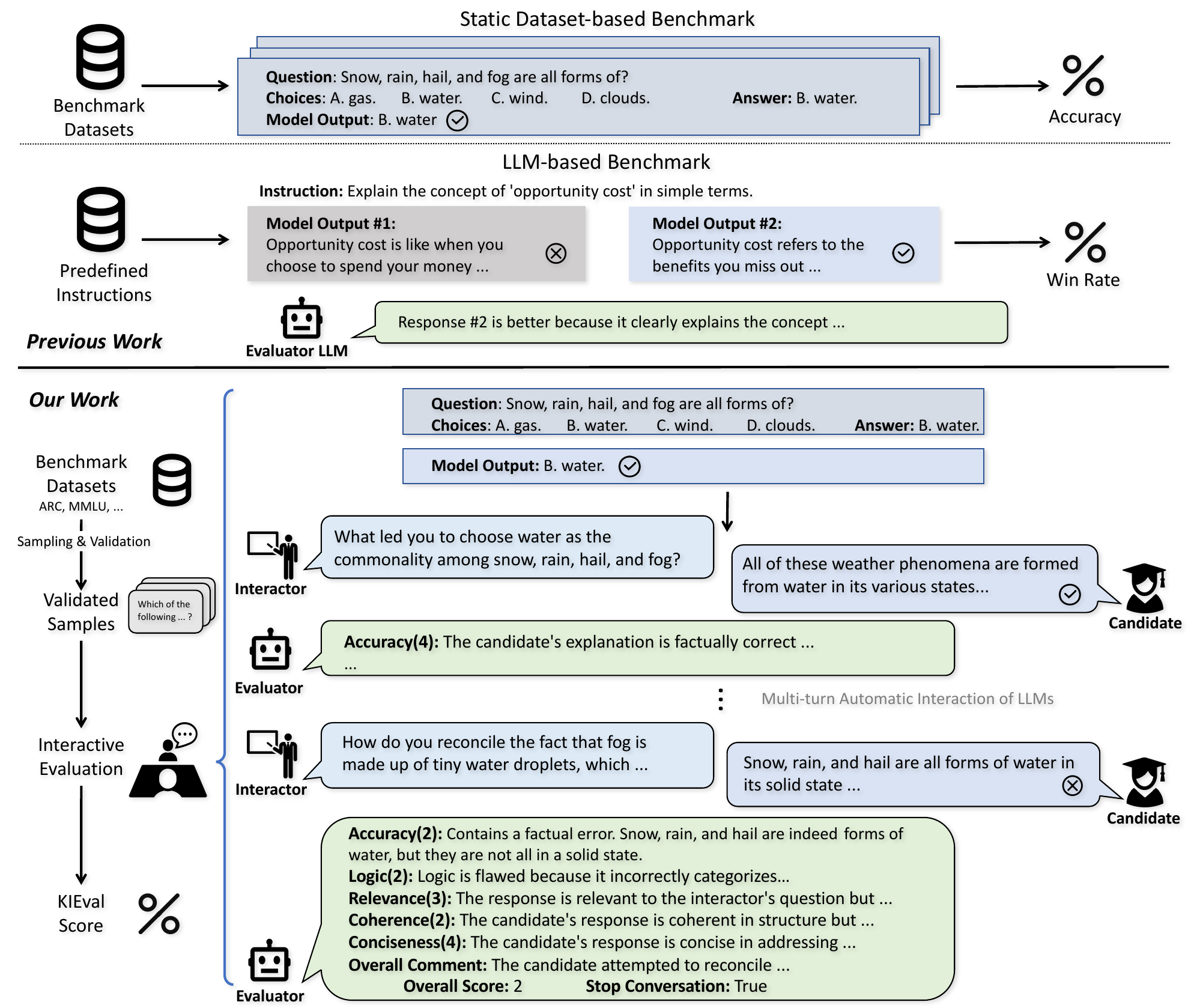}
	\caption{The pipeline of KIEval compared to previous static dataset-based and LLM-based evaluation methods.}
	\label{fig-pipeline}
\end{figure*}

\textbf{Static dataset-based approaches} assess LLMs focused on domain-specific questions or tasks using pre-defined static datasets. Typical evaluation tasks include solving single or multiple-choice problems~\citep{clark2018arc,hendrycks2020mmlu, huang2023ceval} and question answering~\citep{lin2021truthfulqa,cobbe2021traininggsm8k}, these tasks require LLMs to generate short spans of text containing answers to the questions~\citep{eval-harness}. The performance of LLMs is measured by their ability to correctly answer or perform these tasks.

\textbf{LLM-based evaluation}, utilizing one strong LLM~\citep{brown2020language,openai2023gpt4} to assess others, is a recent approach that often employs pairwise comparisons to identify nuanced differences in model outputs, addressing the challenge of determining clear model superiority \citep{pandalm,mtbench}. This method bridges the gap between human and dataset-based evaluations by focusing on generative abilities.
However, this approach has limitations, including reliance on fixed templates~\citep{mtbench}, instructions~\citep{pandalm, alpacaeval}, or multi-round chat datasets~\citep{fu2023gptscore, lin2023llmeval}, limiting its scope in capturing diverse domain knowledge and real-world applicability. It also faces contamination risks, as training on outputs from a strong LLM can inflate results, as noted in work from~\citet{daniele2023amplify-instruct} collect data from MT-Bench~\citep{mtbench} as training data while AlpacaEval~\citep{alpacaeval} contains evaluation set from various instruction-tuning dataset. Additionally, studies indicate these LLM evaluators might be biased~\citep{zeng2023evaluating, wang2023large, pandalm}. While leveraging LLMs to evaluate themselves can be an efficient alternative to human evaluation, understanding and mitigating the potential bias is a crucial problem.

\subsection{Addressing Data Contamination of LLMs}
Data contamination refers to the inclusion of information in the training set of models that provides insights into the test set of a benchmark dataset, and then evaluated in the same benchmark.
Recently, the AI community has become increasingly concerned~\citep{schaeffer2023pretraining, zhou2023dont, oren2023proving} about data contamination in LLMs.
Detecting data contamination, a form of Membership Inference Attack (MIA), poses challenges for large language models (LLMs) due to their training on vast corpora and the difficulty of conducting ablation studies~\citep{shi2023detecting}.
To detect such contamination of LLMs,~\citet{wei2023skywork} suggested comparing average loss values between training and test datasets, while~\citet{shi2023detecting} introduced Min-K\% Prob based on loss values to identify texts used in training. Our experiments show these methods are effective for pre-training but not for detecting contamination during fine-tuning.
\citet{zhu2023dyval} leveraged DAG to dynamically generate evaluation data in reasoning tasks to avoid contamination.
In comparison, KIEval only requires access to output texts of evaluated models and detects data contamination through \emph{evaluating its ability to generalize and utilize knowledge as well as generative ability, which requires a deeper understanding of knowledge instead of mere memorization of the answers}. Moreover, our experiments suggest that KIEval is resilient to data contamination, offering a reliable means to discern whether models have been trained on test sets. This makes it a valuable tool for complementing traditional benchmarks, providing a more nuanced understanding of a model's exposure to and handling of contaminated data.

\section{Methodology}

\subsection{Overview of the KIEval Framework}
KIEval involves a series of iterative interactions, as depicted in Figure \ref{fig-pipeline}.
KIEval is engineered to dynamically evaluate the conversational abilities of LLMs through interactive dialogues focusing on domain-specific topics that challenge LLMs' generative ability and in-depth generalization of knowledge. It simulates realistic conversation flows, offering a dynamic alternative to the static question-answer format of traditional benchmarks.

KIEval orchestrates an evaluation where an LLM, referred to as the \emph{candidate} (the model under evaluation), must understand and respond to an evolving series of questions. These question prompts are generated by an \emph{interactor} model, designed to challenge the candidate with contextually rich scenarios. The responses from the candidate are then assessed by an \emph{evaluator} model, which scrutinizes the output for factual accuracy, relevance, and coherence. The interactor and evaluator are both strong LLMs (e.g., GPT-4, Gemini, Claude 2, LLaMA2-70B-chat, etc.) as the standard practice of LLM-based evaluation protocols.

The design of KIEval emphasizes the importance of reproducibility and consistency in LLM evaluations. By employing separate models for the interactor and evaluator roles, KIEval ensures that the dialogue context remains consistent across different evaluations, as it is fair for the same conversation to be assessed by various evaluators or the same evaluator with different seeds, facilitating a voting strategy to ensure consistent evaluation results. To achieve reproducibility, KIEval utilizes deterministic outputs from LLMs, such as the latest GPT-4 model with temperature sampling disabled and a fixed seed or deploying local models as evaluators. This guarantees identical responses in every run.
Due to space limits, we show the complete system prompts in Appendix~\ref{appendix:complete_prompts}.

\subsection{Interactive Evaluation Procedure}

\begin{algorithm}[t!]
	\caption{\footnotesize KIEval Interactive Evaluation Procedure}
	\label{alg:KIEval}
	{\footnotesize
		\begin{algorithmic}[1]

			\setlength{\itemsep}{1pt}  
			\setlength{\parsep}{0pt}   
			\setlength{\topsep}{1pt}   
			\setlength{\partopsep}{0pt} 

			\REQUIRE Benchmark dataset $\mathcal{Q}$, Interactor model $\mathcal{M}_I$, Candidate model $\mathcal{M}_C$, Evaluator model $\mathcal{M}_E$, seed $r$.
			\STATE Seed everything with $r$, disable temperature sampling for $\mathcal{M}_I$, $\mathcal{M}_C$, $\mathcal{M}_E$ to ensure deterministic outputs.
			\STATE $\mathcal{Q}_S \gets$ Sample subset from $\mathcal{Q}$ with random seed $r$.

			\STATE $\mathcal{Q}_V \gets$ Verify, filter samples from $\mathcal{Q}_S$ with $\mathcal{M}_I$, $\mathcal{M}_E$.

			\FOR{each question $q : (q_{input}, q_{ans})$ in $\mathcal{Q}_V$}
			\STATE Initialize interaction history $S \gets \emptyset$ and evaluation history $E \gets \emptyset$.
			\STATE $q_{pred} \gets$ Predict with $\mathcal{M}_C$ given question $q_{input}$.
			\STATE $\mathcal{O}_{I} \gets$ Generate initial question prompt from $\mathcal{M}_I$ using question $q$ and candidate's answer $q_{pred}$.
			\STATE $S \gets S \cup \{\mathcal{O}_{I}\}$

			\WHILE{not end of dialogue}
			\STATE $\mathcal{O}_{C} \gets$ Generate response from $\mathcal{M}_C$ using $S$.
			\STATE $S \gets S \cup \{\mathcal{O}_{C}\}$.
			\STATE $\mathcal{O}_{E} \gets$ Evaluate response using $\mathcal{M}_E$ with $S$, $E$.
			\STATE $E \gets E \cup \{\mathcal{O}_{E}\}$.
			\IF{Early stopping criteria met for $\mathcal{O}_{C}$}
			\STATE \textbf{break}
			\ENDIF
			\STATE $\mathcal{O}_{I} \gets$ Generate next question from $\mathcal{M}_I$ using $S$.
			\STATE $S \gets S \cup \{\mathcal{O}_{I}\}$
			\ENDWHILE

			\STATE Parse and store results from $E$.
			\ENDFOR
			\STATE $K \gets$ Calculate KIEval scores with $E$.
			\RETURN $K$

		\end{algorithmic}
	}
\end{algorithm}

The interactive evaluation procedure can be described by Algorithm~\ref{alg:KIEval} and the complete implementation can be found in our repository. In LLM-based benchmarks, we hypothesize that the evaluator (\( \mathcal{M}_E \)) models, given their advanced capabilities, can reliably evaluate the performance of less sophisticated candidate models (\( \mathcal{M}_C \))~\citep{mtbench, zeng2023evaluating}. Nevertheless, their applicability as definitive standards is not without limitations, especially when confronting arduous benchmarks. To counteract this, we test the evaluator models against benchmark datasets and sample a fixed number of questions they answer correctly, to ensure the validity of their judgments.

\begin{table*}[!htbp]
	\caption{Evaluation Metrics and Scoring Guide for KIEval. We compute KIEval Score for each metric and a overall KIEval Score as described in~\ref{subsec:eval-metrics}. We design the criteria following \citet{mtbench,pandalm,guo2023evaluating}. }
	\label{tab:combined-metrics-and-guides}
	\resizebox{\textwidth}{!}{
		\centering
		\scriptsize 
		\begin{tabular}{p{0.055\textwidth}p{0.58\textwidth}|p{0.09\textwidth}p{0.3\textwidth}}
			\toprule
			\multicolumn{2}{c|}{{\textbf{Evaluation Metrics}}} & \multicolumn{2}{c}{\textbf{Scoring Guide}}                                                                                                                                                               \\
			\textbf{Metric}                           & \textbf{Description}                                                                                          & \textbf{Score} & \textbf{Criteria}                                              \\
			\midrule
			Accuracy                                  & \scriptsize{Truthfulness and factual correctness of the candidate's response.}                                & 1~Poor         & \scriptsize{Significant deficiencies or inaccuracies.}         \\
			Logic                                     & \scriptsize{Logical structure and soundness of reasoning, including the support and validity of conclusions.} & 2~Below Avg.   & \scriptsize{Noticeable weaknesses, lacking in several areas.}  \\
			Relevance                                 & \scriptsize{The extent to which the response stays on topic and within the scope of the assistant role.}      & 3~Above Avg.   & \scriptsize{Mostly on target with a few minor shortcomings.}   \\
			Coherence                                 & \scriptsize{Integration into the context, consistency with previous statements and conversational flow.}      & 4~Strong       & \scriptsize{Strong performance, often surpasses expectations.} \\
			Conciseness                               & \scriptsize{Brevity and clarity of the response, avoiding unnecessary elaboration or repetition.}             &                &                                                                \\
			\bottomrule
		\end{tabular}
	}
\end{table*}

\subsection{Evaluation Metrics}
\label{subsec:eval-metrics}
KIEval implements a scoring system to quantitatively grade the performance of candidate LLMs in different aspects. Responses are rated on a definitive scale from 1 to 4 for each aspect, where 1 and 4 denote `Poor' and `Strong' performance, respectively, as detailed in \tablename~\ref{tab:combined-metrics-and-guides}. These scores are intended to be definitive to encourage decisive evaluations and are accompanied by comments for interpretability and insights into each score.

We then calculate the KIEval score, which quantitatively measures the results given by the evaluator model, emphasizing sustained and high-quality conversations. Formally, the KIEval score for single-turn scores $s_0, s_1, \ldots, s_{n}$ in $n$ rounds can be computed as:
\begin{equation}
	\notag
	\label{eq:kieval-score}
	\text{KIEvalScore} = \frac{\sum_{i=1}^{n} s_{i} w_{i}}{\sum_{i=1}^{n} w_{i}},
\end{equation}
where the decaying weight $w_i=\exp(-\frac{i}{n})$ placing more emphasis on early turns of the conversation. The normalization ensures a bounded KIEval score, with $1.0$ indicating perfect performance across all rounds.
In addition to these metrics, KIEval incorporates an early stopping mechanism within the evaluative process. The evaluator model (\( \mathcal{M}_E \)) possesses the discretion to prematurely end the conversation if the candidate's response is egregiously inadequate. Criteria for early termination include significant deviations from the topic, empty responses, unpermitted role shifts, and hallucinatory content. We adopt this strategy to measure how well the candidates maintain a meaningful conversation.
\emph{We further examine the effectiveness of these techniques through an ablation study, with detailed experiments and results available in Appendix~\ref{appendix:ablation}. }
\begin{table*}
	\caption{Comparative Evaluation of LLMs using AlpacaEval, MT-Bench and KIEval on different benchmark datasets. We report AlpacaEval win-rates and MT-Bench scores; `Acc.' denotes 5-shot accuracy setting on each dataset or average accuracies in `Overall'; `KIEval' and `Rnds' denote the KIEval score and average rounds of valid conversation rounds.}
	\label{tab:main-results}
	\resizebox{\textwidth}{!}{
		\small
		\setlength{\tabcolsep}{2pt} 
		\renewcommand{\arraystretch}{1.05} 

		\begin{tabular}{l|ccc|ccc|ccc|ccc|ccc|cccccccc}
			\toprule
			                                          &
			\multicolumn{3}{c|}{\small \textbf{ARC-Easy}}      &
			\multicolumn{3}{c|}{\small \textbf{ARC-Challenge}} &
			\multicolumn{3}{c|}{\small \textbf{MMLU}}          &
			\multicolumn{3}{c|}{\small \textbf{HellaSwag}}     &
			\multicolumn{3}{c|}{\small \textbf{C-Eval}}        &
			\multicolumn{4}{c}{\small \textbf{Overall}}          \\
			\multicolumn{1}{c|}{\small }              &
			\scriptsize Acc.                          &
			\scriptsize \makecell{KIEval}             &
			\scriptsize Rnds.                         &
			\scriptsize Acc.                          &
			\scriptsize KIEval                        &
			\scriptsize Rnds.                         &
			\scriptsize Acc.                          &
			\scriptsize KIEval                        &
			\scriptsize Rnds.                         &
			\scriptsize Acc.                          &
			\scriptsize KIEval                        &
			\scriptsize Rnds.                         &
			\scriptsize Acc.                          &
			\scriptsize KIEval                        &
			\scriptsize Rnds.                         &
			\scriptsize Acc.                          &
			\tiny \makecell{AlpacaEval}               &
			\tiny MT-Bench                            &
			\tiny KIEval                                \\ \midrule
			GPT-3.5                                   &
			92.7                                      &
			97.6                                      &
			4.97                                      &
			82.3                                      &
			95.5                                      &
			4.94                                      &
			58.2                                      &
			96.2                                      &
			4.95                                      &
			76.6                                      &
			88.2                                      &
			4.82                                      &
			50.8                                      &
			83.3                                      &
			4.72                                      &
			72.1                                      &
			81.7                                      &
			8.39                                      &
			92.1                                        \\
			LLaMA2 70B                                &
			92.3                                      &
			90.7                                      &
			4.85                                      &
			80.4                                      &
			84.1                                      &
			4.66                                      &
			61.8                                      &
			89.6                                      &
			4.80                                      &
			74.4                                      &
			80.1                                      &
			4.41                                      &
			42.0                                      &
			61.0                                      &
			3.94                                      &
			70.2                                      &
			92.7                                      &
			6.86                                      &
			81.1                                        \\
			LLaMA2 13B                                &
			81.9                                      &
			86.2                                      &
			4.70                                      &
			65.7                                      &
			78.6                                      &
			4.56                                      &
			52.1                                      &
			87.4                                      &
			4.76                                      &
			59.3                                      &
			78.5                                      &
			4.66                                      &
			37.8                                      &
			54.4                                      &
			3.74                                      &
			59.4                                      &
			81.1                                      &
			6.65                                      &
			77.0                                        \\
			LLaMA2 7B                                 &
			73.6                                      &
			78.9                                      &
			4.49                                      &
			55.7                                      &
			74.4                                      &
			4.44                                      &
			44.5                                      &
			83.0                                      &
			4.61                                      &
			39.8                                      &
			76.4                                      &
			4.54                                      &
			33.4                                      &
			49.3                                      &
			3.62                                      &
			49.4                                      &
			71.4                                      &
			6.27                                      &
			72.4                                        \\
			Mistral 7B                                &
			83.5                                      &
			80.8                                      &
			4.64                                      &
			67.5                                      &
			78.5                                      &
			4.46                                      &
			52.7                                      &
			83.0                                      &
			4.62                                      &
			54.4                                      &
			70.3                                      &
			4.34                                      &
			39.3                                      &
			52.2                                      &
			3.61                                      &
			59.5                                      &
			65.5                                      &
			6.84                                      &
			73.0                                        \\
			Yi 6B                                     &
			90.7                                      &
			83.8                                      &
			4.58                                      &
			79.0                                      &
			76.8                                      &
			4.33                                      &
			61.9                                      &
			86.5                                      &
			4.58                                      &
			73.7                                      &
			68.7                                      &
			4.20                                      &
			71.5                                      &
			55.6                                      &
			3.66                                      &
			75.4                                      &
			54.5                                      &
			4.86                                      &
			74.3                                        \\
			MPT 7B                                    &
			53.3                                      &
			68.4                                      &
			4.34                                      &
			43.4                                      &
			65.5                                      &
			4.33                                      &
			33.9                                      &
			74.7                                      &
			4.46                                      &
			27.3                                      &
			57.3                                      &
			4.10                                      &
			26.2                                      &
			44.9                                      &
			3.52                                      &
			36.8                                      &
			43.4                                      &
			5.42                                      &
			62.2                                        \\ \bottomrule
		\end{tabular}
	}
	\renewcommand{\arraystretch}{1.0}
\end{table*}

\begin{figure*}[t!]
	\centering
	\includegraphics[width=1.0\textwidth]{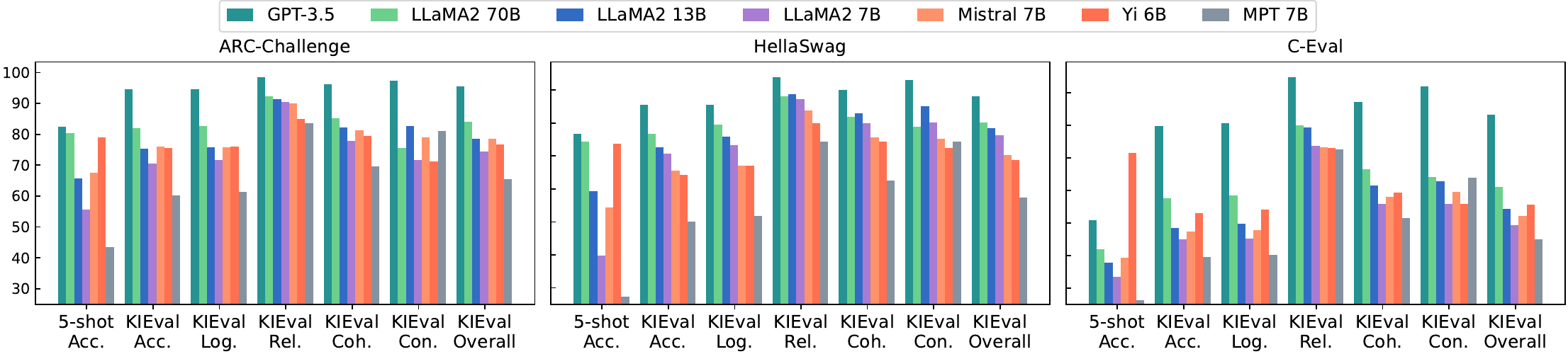}
	\caption{Detailed evaluation result using KIEval, including the overall KIEval score, and KIEval scores for aspects: Accuracy, Logic, Relevance, Coherence and Conciseness. In comparison, we also provide dataset accuracies (5-shot). Due to page limits and the large volume of experimental data, the complete results are put in Appendix ~\ref{appendix:complete_experiments}.}
	\label{fig-dset-details}
\end{figure*}

\section{Experiments}

In this section, we conduct experiments designed to rigorously test the KIEval framework. Our objectives are threefold: (1) to evaluate the generative performance and generalizable knowledge of popular large language models on KIEval using existing benchmark datasets; (2) to assess the impact of data contamination on model performance, specifically examining whether such contamination leads to mere memorization or contributes to genuine understanding and generalization; and (3) to determine the alignment with human, reliability, and effectiveness of KIEval.

\textbf{Experiment Setup.} We select GPT-4\footnote{We use \texttt{gpt-4-1106-preview} from OpenAI's official API for all experiments, including MT-Bench (0.2.32) and AlpacaEval (0.3.6).}~\citep{openai2023gpt4} to be both the evaluator and interactor model by feeding it corresponding prompts with a fixed seed to ensure deterministic outputs. We select 200 samples for each dataset, allowing a maximum of 5 rounds of conversation. The candidates' performance is assessed using the KIEval framework, which evaluates responses based on accuracy, logic, relevance, coherence, and conciseness. We also report dataset-based benchmark accuracies in 5-shot settings and LLM-based benchmark scores from AlpacaEval~\citep{alpacaeval} and MT-Bench~\citep{mtbench} in comparison, as depicted in Table~\ref{tab:main-results}.

\subsection{Evaluation of Popular LLMs by KIEval}

In this experiment, we utilized five popular LLM benchmark datasets: ARC-Easy and ARC-Challenge ~\citep{clark2018arc}, HellaSwag~\citep{zellers2019hellaswag}, MMLU~\citep{hendrycks2020mmlu}, and C-Eval~\citep{huang2023ceval}. For candidate models, we selected a diverse set of 7 LLMs: including proprietary model GPT-3.5~\citep{brown2020language} with API access and open-access foundation models: Llama 2~\citep{touvron2023llama2} 7B, 13B, 70B; Mistral-7B~\citep{jiang2023mistral}; Yi-6B-chat~\citep{yi6b}; MPT-7B~\citep{mpt7b}.\footnote{By default, we use the `chat' versions of Llama2, Yi, and MPT models and the `Instruct' version of Mistral model. Links to the models are released in our GitHub repository.} Detailed introduction of these datasets and models can be found in Appendix~\ref{appendix:appendix_datasets_and_models}.

Referencing Table~\ref{tab:main-results}, we observe the following trends: GPT-3.5 demonstrated consistently high performance across all datasets, particularly excelling in KIEval scores, which indicates strong contextual understanding and response generation. LLaMA2 70B showed competitive results, achieving only a marginal gap from GPT-3.5 on ARC-E, ARC-C, HSwag and even surpasses GPT in MMLU when measured by dataset accuracies, but we can significantly observe a larger gap between these two models with KIEval metrics in all datasets which is also observed by MT-Bench results as reported in Table~\ref{tab:main-results}. This suggests that \emph{traditional benchmarks may underestimate the difference in performance between LLMs as these benchmarks only let models generate a short span of text to evaluate which focus on testing understanding ability. Thus it is hard for these benchmarks to accurately reflect performance gaps in generative tasks.}

The results from different aspects visualized in Figure~\ref{fig-dset-details} benefits us in evaluating model capabilities more comprehensively.
We observe that most models exhibit relatively strong performance in terms of relevance and could generate coherent responses.
Larger models generally perform better in benchmarks, but it is notable that LLaMA2 70B does not perform well in generating concise responses, compared to its smaller counterparts. Although MPT performs weakly in accuracy, its ability to generate concise responses deserves a closer look at its instruction-tuning data.

One interesting finding is that Yi-6B performs unexpectedly well in all benchmark dataset accuracies, especially with it surpasses GPT-3.5 and all other models by a large margin of over 20\% in the C-Eval dataset while exhibiting a similar performance of LLaMA2 70B in other datasets. However, Yi-6B's KIEval score is very similar to LLaMA2 7B and in the range of other 7B models, while it only performs marginally better in the Chinese dataset C-Eval. This raises our concern over potential data contamination in Yi-6B.

To better understand the correlation of KIEval and static dataset-based benchmarks, we provide a detailed analysis in Appendix~\ref{appendix:correlation}.

\subsection{Resilience to Data Contamination}

\begin{table*}
	\caption{Comparison on different data contamination scenarios on ARC-Challenge~\citep{clark2018arc} dataset. `SFT-Cheater' and `PT-Cheater' denote leaking test-set labels during supervised fine-tuning phase and pre-training phase. We report 5-shot accuracy on ARC-Challenge dataset and KIEval scores. We detect data contamination with differences in average language modelling loss~\citep{wei2023skywork} and Min-K\% Prob~\citep{shi2023detecting}.}
	\label{tab:contamination-compare}
	\resizebox{\textwidth}{!}{
		\tiny
		\setlength{\tabcolsep}{2pt} 
		\renewcommand{\arraystretch}{0.95} 

		\begin{tabular}{l|c|ccc|c|cccccc}
			\toprule
			                                & \scriptsize \textbf{ARC-C} & \multicolumn{3}{c|}{\scriptsize \textbf{Avg. LM Loss}} & \scriptsize \textbf{Min-K\%} & \multicolumn{6}{c}{\scriptsize \textbf{KIEval}}                                                  \\
			\multicolumn{1}{c|}{\textbf{}}  &
			{\tiny Acc.(5-shot)}            &
			{\tiny \(\mathcal{L}_{train}\)} &
			{\tiny \(\mathcal{L}_{test}\)}  &
			{\tiny \(\Delta\)}              &
			{\tiny AUC}                     &
			{\tiny Acc.}                    &
			{\tiny Log.}                    &
			{\tiny Rel.}                    &
			{\tiny Coh.}                    &
			{\tiny Con.}                    &
			{\tiny Overall}                                                                                                                                                                                                                                         \\ \midrule
			Normal~(LLaMA 2 7B + SFT)         & 52.8                       & 3.12                                                   & 3.10                         & -0.02                                           & 0.53 & 61.7 & 62.1 & 84.4 & 69.2 & 70.6 & 66.3 \\
			\quad SFT-Cheater               & 69.8                       & 4.05                                                   & 3.95                         & -0.09                                           & 0.54 & 52.8 & 52.3 & 72.8 & 60.2 & 57.7 & 56.1 \\
			\quad PT-Cheater                & 76.8                       & 3.88                                                   & 2.02                         & -1.86                                           & 0.89 & 50.8 & 49.9 & 65.6 & 54.5 & 49.0 & 51.2 \\
			LLaMA 2 7B Chat                  & 57.8                       & 3.05                                                   & 3.01                         & -0.04                                           & 0.55 & 75.3 & 75.9 & 90.1 & 80.2 & 74.0 & 77.9 \\ \bottomrule
		\end{tabular}

	}
	\renewcommand{\arraystretch}{1.0}
\end{table*}

In this subsection, we show that existing static dataset-based and LLM-based evaluation approaches are prone to data contamination while KIEval is resilient to data contamination. Additionally, we test existing contamination detection methods and point out their challenges.

\begin{figure}
	\centering
	\includegraphics[width=0.48\textwidth]{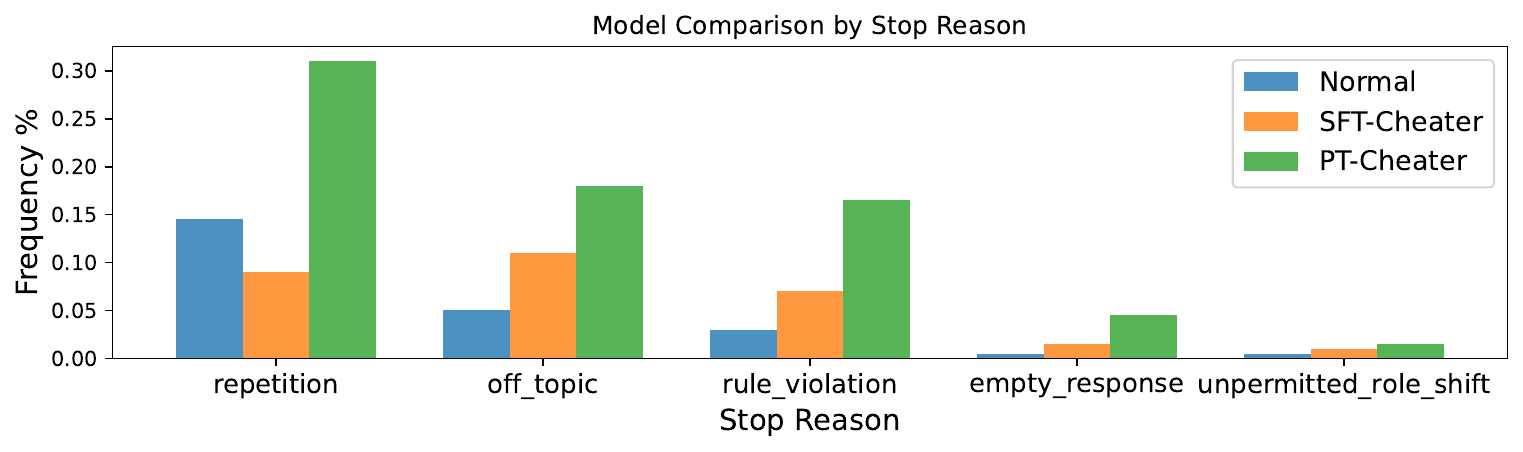}
	\caption{Statistics on reasons to trigger early stopping given by the evaluator model. }
	\label{fig:stop-reasons}
\end{figure}

\textbf{Contamination on static dataset-based evaluation.}
We train two models on the test sets to introduce contamination in the pre-training (`PT-Cheater') and supervised fine-tuning (`SFT-Cheater') phases using un-tuned LLaMA-2 7B as the backbone\footnote{Training details including hyperparameters and hardware settings can be found in Appendix~\ref{appendix:experiment_details}. We also release the full training scripts on our GitHub repository for better reproducibility.}. For PT-Cheater, test set contents are integrated into the pre-training set. Subsequently, the model undergoes fine-tuning with the ShareGPT~\citep{sharegpt}, a commonly used instruction-tuning dataset, to develop chat functionalities. Conversely, the SFT-Cheater replicates this process but adapts the test data to the SFT format. As a control, we also train the backbone solely with ShareGPT (`Normal'), devoid of contamination, ensuring uniform training conditions across all models.
From results in Table~\ref{tab:contamination-compare}, it is clear that the accuracies for benchmarks are significantly boosted, by a large margin of over 45\%, suggesting a susceptibility to data contamination. However, when faced with KIEval, the cheater models perform slightly worse than `Normal' model, not positively affected by data contamination. The average rounds of valid conversation are lower in the cheater models, from the reasons specified by Figure~\ref{fig:stop-reasons}, contaminated models tend to go off-topic of the conversation, repetitively stick to the incorrect knowledge making the conversation meaningless to continue. We can infer from this result that \emph{training models on test sets does not bring generalizable domain knowledge, instead, only contributing to mere memorization of knowledge from test sets.}

\begin{table}
	\caption{Contamination in MT-Bench~\citep{mtbench} scores. We report 5-shot accuracy on ARC-Challenge and KIEval results in comparison.}
	\centering
	\begin{adjustbox}{width=0.3\textwidth}
		\setlength{\tabcolsep}{5pt} 
		\renewcommand{\arraystretch}{0.95} 

		\label{tab:mtbench-contamination}
		\begin{tabular}{lccc}
			\toprule
			\textbf{Model}     & \textbf{Acc.}  & \textbf{MT-Bench} & \textbf{KIEval} \\ \midrule
			Normal    & 52.35 & 3.96     & 62.60  \\
			+MT-Bench & 52.25 & 5.75     & 57.46  \\ \bottomrule
		\end{tabular}
	\end{adjustbox}
	\renewcommand{\arraystretch}{1.0} 

\end{table}

\textbf{Contamination on LLM-based evaluation.}
We also find existing LLM-based evaluations vulnerable to data contamination, due to their reliance on static templates. We train the fine-tuned model (`Normal') with MT-Bench input templates and GPT-4 outputs using only 80 samples and test it against MT-Bench and KIEval.
Table~\ref{tab:mtbench-contamination} reveals that contamination training notably inflates the MT-Bench score by 1.79, a surge over 45\% compared to the baseline, while ARC-Challenge accuracy remains stable and KIEval score slightly decreased.

\textbf{Challenges in Contamination Detection.}
We evaluate the efficacy of current data contamination detection strategies, notably Skywork~\citep{wei2023skywork} and Min-K\% Prob~\citep{shi2023detecting}, which identify training data leakage through loss metrics as introduced in Related Work. We sampled 200 instances each from the trainset and testset of ARC-Challenge with contamination labels and tried to classify each instance with Min-K\% Prob. We report AUC to measure its effectiveness.
Table~\ref{tab:contamination-compare} demonstrates their capability to detect leaked data in the pre-training phase effectively, as the difference of average loss is significantly higher and Min-K\% AUC reaches 0.89. However, both methods fail to identify contamination during SFT, with slight differences in loss values and Min-K\% Prob AUC near random.
We hypothesize that this may be due to the fine-tuning process only supervising the output sequence which is a short span containing the answer. This enables easy recall of the answer, without significantly impacting average loss values or Min-K\% Prob values.
\emph{This discrepancy underscores the ineffectiveness of loss-based metrics in discerning data contamination during SFT phase.} Conversely, by leveraging KIEval, a correlation between KIEval scores and dataset accuracies emerges, suggesting the potential of KIEval in distinguishing between generalized knowledge application and mere data regurgitation for contamination detection.

\subsection{Meta-Evaluation of KIEval}

Meta-evaluation is essential for validating the practical utility of any evaluation framework. In this section, we assess KIEval's alignment with human judgment and compare its performance against existing evaluation methods. Additionally, we conduct a cost analysis focusing on computational resources and API usage in Appendix~\ref{appendix:cost} to validate KIEval's cost-effectiveness and scalability.

\begin{table}
	\caption{Pearson (\(r\)), Spearman (\(\rho\)), Kendall-Tau (\(\tau\)) correlation efficients of METEOR~\citep{banerjee2005meteor}, ROUGE~\citep{lin2004rouge}, BERTScore~\citep{zhang2019bertscore}, and MT-Bench~\citep{mtbench} with human scores and their variance (\(\sigma^2\)). We introduce the compared baselines in Appendix~\ref{appendix:alignment_baselines}, ~(\(\uparrow\)) indicates higher values are better.}
	\centering
	\begin{adjustbox}{width=0.36\textwidth}
		\label{tab:alignment}
		\setlength{\tabcolsep}{6pt} 
		\renewcommand{\arraystretch}{0.95} 
		\begin{tabular}{l|llll}
			\toprule
			\multicolumn{1}{c|}{\textbf{Metric}}             &
			\multicolumn{1}{c}{\textbf{\(r~(\uparrow\))} }   &
			\multicolumn{1}{c}{\textbf{\(\rho~(\uparrow\))}} &
			\multicolumn{1}{c}{\textbf{\(\tau~(\uparrow\))}} &
			\multicolumn{1}{c}{\textbf{\(\sigma^2\)}}                                        \\ \midrule
			METEOR                                           & 0.016 & 0.023 & 0.021 & 0.012 \\
			ROUGE-1                                          & 0.259 & 0.316 & 0.226 & 0.016 \\
			ROUGE-2                                          & 0.280 & 0.303 & 0.223 & 0.007 \\
			ROUGE-L                                          & 0.209 & 0.268 & 0.200 & 0.007 \\
			BERTScore                                        & 0.189 & 0.336 & 0.250 & 0.001 \\
			MT-Bench                                         & 0.520 & 0.494 & 0.405 & 9.360 \\ \midrule
			\textbf{Ours}                                    &       &       &       &       \\
			Accuracy                                         & 0.761 & 0.727 & 0.653 & 2.010 \\
			Logic                                            & 0.768 & 0.735 & 0.661 & 1.842 \\
			Relevance                                        & 0.633 & 0.643 & 0.543 & 1.152 \\
			Coherence                                        & 0.750 & 0.740 & 0.644 & 1.365 \\
			Conciseness                                      & 0.611 & 0.604 & 0.504 & 0.833 \\
			Overall                                          & 0.814 & 0.789 & 0.721 & 1.512 \\ \bottomrule
		\end{tabular}
		\renewcommand{\arraystretch}{1.0}

	\end{adjustbox}
\end{table}

\textbf{Human Evaluation:}
To ascertain KIEval's alignment with human preferences and its comparative effectiveness against prior methods, we collected a sample of multi-turn conversations generated by KIEval and compared the correlation between different evaluators and human-annotated scores.
Specifically, we sampled 100 sets of conversations from all 5 datasets and across 7 candidate models and converted the multi-turn conversations to single-turn format to evaluate only one round of interaction to align with our compared baselines.
Three human experts were asked to independently rate the responses of different models on a scale from 1 to 4, and we cover human annotation details in Appendix~\ref{appendix:human_annotation}. The Inter-Annotator Agreement (IAA) was measured by averaging Cohen's Kappa coefficients for each annotator pair, yielding an average IAA of 0.624. This indicates substantial agreement among the annotators, a significant achievement for the complexity of the task at hand. The average score for each instance was then calculated and used as the human score for that response.

Following the meta-evaluation in G-Eval \citep{liu2023gpteval}, we computed Pearson, Spearman, and Kendall-Tau correlation coefficients to gauge the agreement between different evaluators' scores and human ratings. A detailed introduction of these evaluated baselines is provided in Appendix~\ref{appendix:alignment_baselines} due to page limitations.

As shown in Table~\ref{tab:alignment}, traditional reference-based evaluators align poorly with human judgments in open-ended conversations, due to their reliance on limited reference texts which cannot encompass the variety of valid responses. While the LLM-based evaluator MT-Bench shows commendable alignment with human preferences, its applicability is somewhat limited by its design, which is tailored to a predefined set of instructions and responses. In contrast, KIEval demonstrates a robust correlation with human preferences, underscoring its efficacy in evaluating dynamically generated, open-ended conversations.

\textbf{Potential Bias:} Employing LLMs for evaluation could introduce additional biases into the evaluation results. To mitigate bias and enhance objectivity, we have designed a separation between the 'interactor' and 'evaluator' roles. By utilizing different LLMs as evaluators with a fixed interactor LLM, we can assess the same conversation (since the interactor and candidate models are fixed) multiple times with different evaluators. Results in Appendix~\ref{appendix:bias} indicate that while different evaluator LLMs may have varying preferences for model outputs, their overall results demonstrate a strong correlation.

\textbf{Ablation Study:} We further examine the effect of KIEval's main components through an ablation study, the experiments and results are presented in Appendix~\ref{appendix:ablation}.

\section{Conclusion}

KIEval provides a dynamic evaluation and analysis of LLMs across various domains, evaluating generative abilities and domain knowledge through structured conversations instead of relying on fixed templates or instructions, reducing the risk of data contamination and enhancing the reliability of evaluations, while preserving alignment with human preference. Overall, our findings suggest several key insights:
\begin{itemize}[leftmargin=0.2em]
	\setlength\itemsep{0em}
	\item Static dataset-based benchmarks may not capture the full extent of performance disparities among LLMs, such datasets could potentially underestimate these differences.
	\item Training models on test splits of benchmark datasets primarily improves recall of answers rather than a genuine enhancement in knowledge comprehension or problem-solving abilities, underscoring the impact of data contamination.
	\item Detecting data contamination, particularly for the fine-tuning phase of LLMs, might be challenging for existing methods. We propose a paradigm shift from only detecting exposure to specific training texts towards evaluating the models' underlying rationale and depth of knowledge comprehension.

\end{itemize}

We believe that KIEval will serve as a valuable tool for researchers and practitioners alike, aiding in the development of more robust, versatile, and ethical AI systems.

\section{Limitations}

Our method, while insightful, operates under the assumption that LLMs can accurately evaluate the capabilities of less sophisticated models. However, the reliability of LLMs as universal evaluators is not without limitations, particularly when faced with complex benchmarks or assessing more advanced models. For certain evaluation tasks, such as mathematics problem-solving, coding, and fact-checking, depending solely on LLM evaluators may be insufficient. Furthermore, these evaluators may introduce additional biases into the assessment process. As these limitations can also be applicable to other current LLM-based evaluators, future research could explore a hybrid evaluation strategy that combines task-specific methods with LLM evaluators to achieve more nuanced and accurate assessments.

Another limitation concerns the scope of our work. Our focus is on evaluating instruction-tuned generative models with conversational abilities, excluding those designed solely for natural language understanding (NLU) tasks without generative capabilities or base models lacking instruction-following capabilities. We can assess base models by instruction-tuning them using the exact same datasets and settings, operating under the hypothesis that employing identical data for training different models results in a fair comparison. Future research should delve more deeply into the evaluation of base models, scrutinizing the impact of instruction-tuning on their performance.

\bibliography{anthology,custom}
\bibliographystyle{acl_natbib}

\clearpage

\appendix

\section{Baseline Evaluators}

In our experimental framework, we compare KIEval with prevalent evaluators in open-ended dialogue evaluation, following~\citet{liu2023gpteval}, alongside MT-Bench, which epitomizes the application of Large Language Models (LLMs) in evaluation processes. To compare reference-based methods with reference-free approaches and our method, we use GPT-4 to generate references.

\label{appendix:alignment_baselines}

\begin{itemize}[leftmargin=1em]
	\setlength\itemsep{0em}
	\item \textbf{METEOR}~\citep{banerjee2005meteor}, a reference-based evaluation metric, utilizes unigram matching between generated outputs and reference texts crafted by humans to assess performance across a variety of Natural Language Generation (NLG) tasks, including machine translation and dialogue generation.

	\item \textbf{ROUGE}~\citep{lin2004rouge} comprises a suite of metrics for reference-based evaluation, facilitating the comparison of automatically generated summaries or translations against one or more human-crafted reference summaries or translations.

	\item \textbf{BERTScore}~\citep{zhang2019bertscore}, another reference-based evaluation metric, employs contextual embeddings from BERT to measure cosine similarity between words in candidate and reference sentences. Demonstrated to align well with human judgment at both the sentence and system levels, BERTScore calculates precision, recall, and F1 scores, offering valuable insights for various NLG tasks.

	\item \textbf{MT-Bench}~\citep{mtbench}, a LLM-based, reference-free evaluation approach, harnesses cutting-edge LLMs to assess model outputs. It features a series of open-ended questions designed to test a model's capabilities in engaging in conversation and instruction-following abilities. As MT-Bench is similar to AlpacaEval~\citep{alpacaeval}, PandaLM~\citep{pandalm} and G-Eval~\citep{liu2023gpteval} but being a more popular option, we select MT-Bench without compromising on the breadth of our evaluation. In our meta-evaluation experiment, we use \texttt{gpt-4-1106-preview} as the evaluator and use the single-answer grading mode of MT-Bench.
\end{itemize}

\section{Datasets}
\label{appendix:appendix_datasets_and_models}

\begin{table*}
	\centering
	\caption{Details of datasets in our experiments. We report 5-shot accuracy metric of `Used Splits' split for each dataset.}
	\label{tab:data-statistics}
	\begin{adjustbox}{width=0.8\textwidth}
		\begin{tabular}{lllll}
			\toprule
			\textbf{Datasets} & \textbf{Splits}                         & \textbf{Used Splits} & \textbf{Split Size} & \textbf{Language} \\ \midrule
			ARC-Challenge     & train, validation, test                 & test                 & 1.17k               & English           \\
			ARC-Easy          & train, validation, test                 & test                 & 2.38k               & English           \\
			Hellaswag         & train, validation,test                  & validation           & 10k                 & English           \\
			MMLU              & auxiliary\_train, test, validation, dev & test                 & 14k                 & English           \\
			C-Eval            & val, test, dev                          & val                  & 1.35k               & Chinese           \\ \bottomrule
		\end{tabular}
	\end{adjustbox}
\end{table*}
We use the following datasets in our experiments, for statistics and used splits, please refer to Table~\ref{tab:data-statistics}.

\textbf{ARC-Easy and ARC-Challenge~\citep{clark2018arc}:} Both are subsets of the AI2 Reasoning Challenge, a benchmark for assessing a model's reasoning and understanding in science questions. ARC-Easy contains simpler questions, while ARC-Challenge includes more complex ones.

\textbf{HellaSwag~\citep{zellers2019hellaswag}:} challenges models to complete realistic scenarios in text, testing common sense and predictive abilities.

\textbf{MMLU~\citep{hendrycks2020mmlu}:} A comprehensive English examination composed of multiple-choice questions encompassing a wide array of disciplines. This extensive test includes subjects ranging from humanities and social sciences to hard sciences, alongside other essential areas of knowledge. It encompasses 57 distinct tasks, covering fields such as elementary mathematics, US history, computer science, law, and beyond.

\textbf{C-Eval~\citep{huang2023ceval}:} A comprehensive Chinese evaluation composed of 13948 multi-choice questions spanning 52 diverse disciplines and four difficulty levels.

\section{Correlation Analysis of KIEval and Dataset Benchmarks}
\label{appendix:correlation}

\begin{table}
	\caption{Pearson correlation coefficient of KIEval scores and dataset accuracy scores. Due to suspected data contamination in Yi-6B, we report two sets of results with and without Yi.}
	\centering
	\begin{adjustbox}{width=0.4\textwidth}
		\setlength{\tabcolsep}{3pt} 
		\renewcommand{\arraystretch}{0.85} 
		\label{tab:pearson}
		\begin{tabular}{ccccc}

			\toprule
			PCC       & $r$   & $p$      & \thead{$r$            \\ Excl. Yi} & \thead{$p$\\ Excl. Yi} \\ \midrule
			Overall   & 0.664 & 1.37E-05 & 0.765      & 8.67E-07 \\
			ARC-E     & 0.892 & 6.97E-03 & 0.934      & 6.45E-03 \\
			ARC-C     & 0.839 & 1.83E-02 & 0.940      & 5.29E-03 \\
			MMLU      & 0.814 & 2.57E-02 & 0.876      & 2.21E-02 \\
			HellaSwag & 0.686 & 8.85E-02 & 0.862      & 2.74E-02 \\
			C-Eval    & 0.427 & 3.40E-01 & 0.924      & 8.42E-03 \\ \bottomrule
		\end{tabular}
	\end{adjustbox}
	\renewcommand{\arraystretch}{1} 

\end{table}

\begin{figure}[t!]
	\centering
	\includegraphics[width=0.48\textwidth]{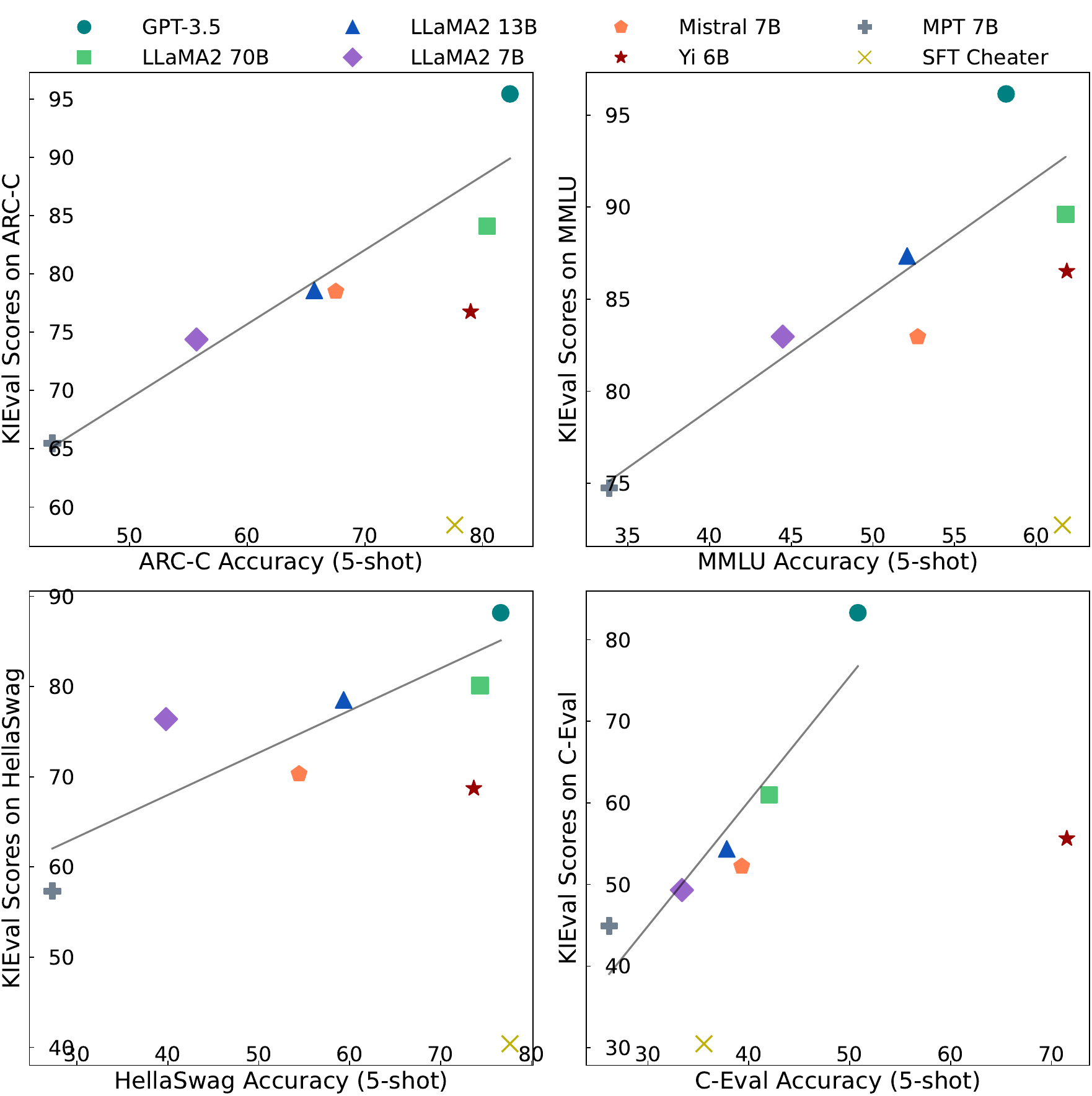}
	\caption{Scatter plots of KIEval scores and traditional benchmark scores by model and dataset. Each point represents the performance of a model on a specific dataset, measured by the KIEval score and accuracy score (5-shot). Regression lines are plotted for each dataset. \emph{Points significantly above the regression line indicate the performance gap not captured by traditional benchmarks but captured by KIEval, while points significantly below the regression line indicate potential data contamination in traditional benchmarks.} }
	\label{fig-regression}
\end{figure}

To further investigate the correlation between dataset-based benchmarks and KIEval, we use regression analysis as shown in Figure~\ref{fig-regression}. We also leverage the Pearson correlation coefficient to provide quantitive analysis in Table~\ref{tab:pearson}.
The results revealed a significant positive correlation between KIEval scores and dataset-based benchmark accuracies. This correlation underscores KIEval's alignment with traditional evaluation methods. However, we also bring new insights that traditional benchmarks do not offer: \emph{while dataset-based benchmarks effectively assess LLM knowledge under contamination-free conditions, their results are easily inflated in the presence of data contamination. In contrast, KIEval exhibits a lower susceptibility to these issues.}
Visual analysis offers additional perspective by contrasting model performances as per benchmark accuracies and KIEval scores. Models significantly above the regression line suggest capabilities beyond those captured by traditional benchmarks. In this scenario, traditional benchmarks are not sufficiently challenging to effectively differentiate the stronger models from others, nor do they accurately represent the generative capabilities of these models. It is evident that GPT-3.5 is included in this category. Conversely, models falling below the regression line, exhibiting high benchmark accuracy but low conversation quality, suggest limited real-world applicability, potentially indicative of data contamination. Interestingly, the visualization shows that not only does our simulated SFT Cheater model fall into the outlier category below the regression line, but Yi-6B also exhibits similar behavior.

\section{Ablation Study of KIEval Components}
\label{appendix:ablation}

\begin{table}
	\caption{Ablation study of KIEval's components. DW and ES denote decaying weight in KIEval score and early stopping mechanism. We report Pearson (\(r\)), Spearman (\(\rho\)), Kendall-Tau (\(\tau\)) correlation efficients with human scores and their variance (\(\sigma^2\)).~(\(\uparrow\)) indicates higher values are better.}
	\centering
	\begin{adjustbox}{width=0.4\textwidth}
		\label{tab:ablation}
		\setlength{\tabcolsep}{4pt} 
		\renewcommand{\arraystretch}{0.9} 
		\begin{tabular}{cc|cccc}
			\toprule
			\multicolumn{2}{c|}{Methods}                  &
			\multirow{2}{*}{\textbf{\(r~(\uparrow\))}}    &
			\multirow{2}{*}{\textbf{\(\rho~(\uparrow\))}} &
			\multirow{2}{*}{\textbf{\(\tau~(\uparrow\))}} &
			\multirow{2}{*}{\textbf{\(\sigma^2\)}}                                                                                \\
			\multicolumn{1}{c}{\footnotesize DW}          & \multicolumn{1}{c|}{\footnotesize ES} &       &       &       &       \\ \midrule
			\cmark                                        & \cmark                                & 0.768 & 0.759 & 0.625 & 1.718 \\
			\xmark                                        & \cmark                                & 0.730 & 0.721 & 0.588 & 1.708 \\
			\cmark                                        & \xmark                                & 0.737 & 0.751 & 0.623 & 1.839 \\
			\xmark                                        & \xmark                                & 0.691 & 0.715 & 0.588 & 1.870 \\ \bottomrule
		\end{tabular}
		\renewcommand{\arraystretch}{1.0}

	\end{adjustbox}
\end{table}

This study assesses the impact of the decaying weight scoring and the early stopping mechanism of KIEval through an ablation analysis. Employing the same set of KIEval-generated conversations used in our meta-evaluation, we explore four distinct configurations of the KIEval framework. Specifically, we investigate the influence of the weighted scoring by replacing the decaying weight with a constant value, effectively equating the multi-round score to the mean of single-turn scores. Additionally, we examine the consequences of omitting the early stopping mechanism, thereby allowing conversations to proceed unabated until their conclusion. We then compare the correlation coefficients between these variants and human scores. As indicated by the data in Table~\ref{tab:ablation}, the exclusion of either feature results in a notable decline in performance, underscoring their respective contributions to the model's efficacy.

\section{Cost and Scalability}
\label{appendix:cost}

Assessing KIEval's scalability requires a thorough evaluation of overall costs. Our method employs a strong LLM accessed via API, with expenses based on input and output token lengths. Table~\ref{tab:token-count-total} details the average token count per model evaluation across diverse datasets. Additionally, the average GPU expenditure for single model evaluations on NVIDIA A100 GPUs is provided in Table~\ref{tab:gpu-hours}. Financially, deploying GPT-4 in both interactor and evaluator roles within KIEval incurs a cost of around 27 USD for each model evaluation, comprising 1000 interaction rounds. Importantly, due to our adoption of single-answer grading over pairwise comparison~\citep{pandalm, mtbench}, costs increase linearly rather than quadratically with the number of models evaluated. For a comprehensive understanding of the cost implications at scale, we present a detailed estimation in Table~\ref{tab:cost-scale}.

\begin{table}
	\caption{API usage estimation for KIEval and pairwise-comparison based evaluation methods. Priced in USD, according to openai's GPT-4 pricing policy.}
	\centering
	\begin{adjustbox}{width=0.35\textwidth}
		\setlength{\tabcolsep}{3pt} 
		\renewcommand{\arraystretch}{0.85} 

		\label{tab:cost-scale}
		\begin{tabular}{ccccl}
			\toprule
			Method   & 1 Model & 10 Models & 100 Models & \\ \midrule
			KIEval   & 27      & 279       & 2,796      & \\
			Pairwise & 16      & 720       & 79,200     & \\ \bottomrule
		\end{tabular}
	\end{adjustbox}
\end{table}
\begin{table}
	\caption{Average GPU budget for a single model evaluated on one dataset with KIEval. We report results for LLaMA2 models with varying parameter sizes.}
	\centering
	\begin{adjustbox}{width=0.28\textwidth}
		\label{tab:gpu-hours}
		\begin{tabular}{cccc}
			\toprule
			                   & {\small 7B}   & {\small 13B}  & {\small 70B}  \\ \midrule
			{\small GPU Hours} & {\small 0.74} & {\small 0.99} & {\small 9.38} \\ \bottomrule
		\end{tabular}
	\end{adjustbox}
\end{table}
\begin{table}
	\caption{Average number of tokens consumed of evaluation on a single model across various datasets, over 200 samples with 5 rounds of interaction.}
	\centering
	\begin{adjustbox}{width=0.38\textwidth}
		\label{tab:token-count-total}
		\begin{tabular}{l|cc|cc}
			\toprule
			                                       &
			\multicolumn{2}{c|}{\small Interactor} &
			\multicolumn{2}{c}{\small Evaluator}                               \\
			\multicolumn{1}{c|}{\small }           &
			\scriptsize Prompt                     &
			\scriptsize \makecell{Completion}      &
			\scriptsize Prompt                     &
			\scriptsize Completion                                             \\ \midrule
			Avg.                                   & 557k & 28k & 1546k & 203k \\
			ARC-E                                  & 554k & 28k & 1592k & 208k \\
			ARC-C                                  & 540k & 27k & 1553k & 205k \\
			MMLU                                   & 656k & 30k & 1731k & 213k \\
			HellaSwag                              & 527k & 29k & 1488k & 198k \\
			C-Eval                                 & 505k & 26k & 1365k & 189k \\ \bottomrule
		\end{tabular}
	\end{adjustbox}
\end{table}

\section{Experiment Details}
\label{appendix:experiment_details}

In this section, we detail the experimental setup employed to facilitate the reproduction of our results. The entire codebase used in our experiments has been made publicly available, ensuring transparency and ease of verification for our findings.

\textbf{Codebase and Dependencies}: Our experiments are done with LLaMA-Factory package\footnote{Available at \url{https://github.com/hiyouga/LLaMA-Factory/}.}, a framework designed to streamline the training of large language models. We use Huggingface Transformers~\citep{wolf2019huggingface} library and the Deepspeed~\citep{rasley2020deepspeed} Zero-3 optimizer~\citep{rajbhandari2021zero}, forms the backbone of our computational experiments.

\textbf{Training Configuration}: For the training process, we have configured the learning rate to 2e-5, employing a cosine learning rate scheduler. Our hardware setup consists of 4 NVIDIA A100 GPUs, and we've set per-device batch size to 1, coupled with a gradient accumulation step of 4. We use full-parameter training for 4 epochs in all our experiments, including training models with data contamination during pre-training and fine-tuning.

\section{Human Annotation Details}
\label{appendix:human_annotation}

\begin{table}
	\centering
	\caption{We report Cohen's Kappa (\(\kappa\)) for each pair of annotators to measure Inter-Annotator Agreement (IAA) between annotators.}
	\begin{adjustbox}{width=0.24\textwidth}
		\scriptsize
		\begin{tabular}{ccc}
			\toprule
			\textbf{Annotator Pair} & \textbf{\(\kappa\)} \\
			\midrule
			A + B                   & 0.650               \\
			A + C                   & 0.580               \\
			B + C                   & 0.642               \\
			\bottomrule
		\end{tabular}
	\end{adjustbox}
	\label{tab:iaa}
\end{table}

For human annotation in our work, all annotators are authors of this paper who previously have not accessed the outputs of models in our experiments and volunteer to contribute. All three annotators agree on how the data would be used. Since the data to be annotated come from open-source datasets and popular LLMs, ethical concerns are not applicable. We provide guides for each annotator and for each annotator, we give them a unique URL to our annotation platform built with Gradio, and give them instructions: `You are given some conversations between a candidate model and a interactor model. Please score the response of the candidate model with integers from 1 to 4, following our scoring guide. Your score should be definitive, and consider the response's factual accuracy, logical structure, language conciseness, and coherence.'

We measure the agreement of our annotators by calculating Cohen's Kappa as Inter-Annotator Agreement, results can be found in Table~\ref{tab:iaa}. We reach an average IAA for all pairs of human annotators of 0.624, indicating a substantial agreement among our annotators.

\section{Potential Bias}
\label{appendix:bias}

While KIEval provides a new evaluation method, the reliance on strong LLMs as evaluators could inadvertently propagate existing biases. To study the bias introduced by the evaluator LLMs, we utilize different LLMs as evaluators with a fixed interactor LLM. Specifically, we use \texttt{gpt-4-turbo-preview-1106} and \texttt{claude-3-opus-20240229} as evaluators, with the same prompts and sampling hyperparameters. We report KIEval scores of \texttt{gpt-3.5-turbo}, \texttt{llama-2-70b-chat-hf} and \texttt{llama-2-7b-chat-hf} from different evaluator LLMs on ARC-Challenge dataset in Table~\ref{tab:bias-detail}. We also report the correlation coefficients of the results in Table~\ref{tab:bias-correlation}.

These results indicate that although GPT-4 and Claude 3 have different preference of models, their overall results exhibit a strong correlation. Note that we use the exact same prompt for GPT-4 and Claude 3, and as Claude 3 works differently with their system prompts, the scores from Claude 3 are generally higher but this does not affect the effectiveness of our experiments.

\begin{table}
\scriptsize
	\caption{KIEval scores using Claude 3 Opus (\texttt{claude-3-opus-20240229}) and GPT-4 Turbo (\texttt{gpt-4-turbo-preview-1106}) as evaluators on ARC-Challenge dataset. }
	\centering
	\begin{adjustbox}{width=0.49\textwidth}
		\label{tab:bias-detail}
		\setlength{\tabcolsep}{2pt} 
		\renewcommand{\arraystretch}{1.1} 
\begin{tabular}{lccccccc}
\toprule
\textbf{\tiny Model} &
  \textbf{\tiny Evaluator} &
  \textbf{\tiny Accuracy} &
  \textbf{\tiny Logic} &
  \textbf{\tiny Relevance} &
  \textbf{\tiny Coherence} &
  \textbf{\tiny Conciseness} &
  \textbf{\tiny Overall}
   \\ \midrule
\multirow{2}{*}{GPT-3.5} &
  GPT-4 &
  94.6 &
  94.7 &
  98.5 &
  96.1 &
  97.3 &
  95.5
   \\
 &
  Claude-3 &
  98.6 &
  98.8 &
  99.8 &
  99.4 &
  99.0 &
  98.7
   \\
\multirow{2}{*}{LLaMA-2 70B} &
  GPT-4 &
  81.9 &
  82.8 &
  92.2 &
  85.3 &
  75.6 &
  84.1
   \\
 &
  Claude-3 &
  98.3 &
  98.7 &
  98.2 &
  96.9 &
  84.6 &
  96.4
   \\
\multirow{2}{*}{LLaMA-2 7B} &
  GPT-4 &
  70.6 &
  71.6 &
  90.4 &
  77.9 &
  71.7 &
  74.4
   \\
 &
  Claude-3 &
  90.9 &
  91.8 &
  98.0 &
  95.0 &
  85.2 &
  91.0 \\ \bottomrule
\end{tabular}
		\renewcommand{\arraystretch}{1.0}

	\end{adjustbox}
\end{table}
\begin{table}
	\caption{Pearson (\(r\)), Spearman (\(\rho\)), Kendall-Tau (\(\tau\)) correlation efficients of KIEval scores evaluated by \texttt{claude-3-opus-20240229} and \texttt{gpt-4-turbo-preview-1106}. }
	\centering
	\begin{adjustbox}{width=0.32\textwidth}
		\label{tab:bias-correlation}
		\setlength{\tabcolsep}{4pt} 
		\renewcommand{\arraystretch}{0.9} 
		\begin{tabular}{lcc}
\toprule
\textbf{Metric}   & \textbf{Corr. Coeff.} & \textbf{P-Value}   \\ \midrule
Pearson \(r\) & 0.822                   & 2.87e-05  \\
Spearman \(\rho\) & 0.898                   & 4.17e-07  \\
Kendall \(\tau\) & 0.761                   & 1.10e-05 \\ \bottomrule
\end{tabular}
		\renewcommand{\arraystretch}{1.0}

	\end{adjustbox}
\end{table}

\section{Use of AI Assistants}

In this work, we use GitHub Copilot to assist coding, and GPT-4 to correct grammatical errors.

\section{Complete Experiment Results}
\label{appendix:complete_experiments}

We share the complete experiment results from all 5 datasets with 7 models, evaluated with KIEval and benchmark accuracies in Table~\ref{tab:arce-details},~\ref{tab:arcc-details},~\ref{tab:mmlu-details},~\ref{tab:hellaswag-details},~\ref{tab:ceval-details}.

\begin{table*}
	\setlength{\tabcolsep}{3pt} 

	\centering
	\caption{KIEval Results on ARC-Easy.}
	\label{tab:arce-details}
	\small
	\begin{tabular}{l|ccccccc|c}
		\toprule
		\multicolumn{1}{c|}{ARC-E} & \scriptsize Accuracy & \scriptsize Logic & \scriptsize Relevance & \scriptsize Coherence & \scriptsize Conciseness & \scriptsize Overall & \scriptsize Rounds & \scriptsize Acc. (5-shot) \\ \midrule
		GPT-3.5                    & 97.1                 & 97.4              & 99.3                  & 97.9                  & 97.9                    & 97.6                & 4.97               & 92.7                      \\
		LLaMA2 70B                 & 90.3                 & 90.3              & 94.6                  & 91.3                  & 79.6                    & 90.7                & 4.85               & 92.3                      \\
		LLaMA2 13B                 & 84.5                 & 84.3              & 93.2                  & 87.7                  & 85.8                    & 86.2                & 4.70               & 81.9                      \\
		LLaMA2 7B                  & 77.1                 & 77.4              & 89.7                  & 82.2                  & 73.6                    & 78.9                & 4.49               & 73.6                      \\
		Mistral 7B                 & 78.5                 & 78.2              & 91.4                  & 83.5                  & 79.9                    & 80.8                & 4.64               & 83.5                      \\
		Yi 6B                      & 83.4                 & 83.6              & 90.9                  & 85.8                  & 76.4                    & 83.8                & 4.58               & 90.7                      \\
		MPT 7B                     & 63.9                 & 64.1              & 84.9                  & 71.5                  & 81.8                    & 68.4                & 4.34               & 53.3                      \\ \bottomrule
	\end{tabular}
\end{table*}

\begin{table*}
	\setlength{\tabcolsep}{3pt} 

	\centering
	\caption{KIEval Results on ARC-Challenge.}
	\label{tab:arcc-details}
	\small
	\begin{tabular}{l|ccccccc|c}
		\toprule
		\multicolumn{1}{c|}{ARC-C} & \scriptsize Accuracy & \scriptsize Logic & \scriptsize Relevance & \scriptsize Coherence & \scriptsize Conciseness & \scriptsize Overall & \scriptsize Rounds & \scriptsize Acc. (5-shot) \\ \midrule
		GPT-3.5                    & 94.6                 & 94.7              & 98.5                  & 96.1                  & 97.3                    & 95.5                & 4.94               & 82.3                      \\
		LLaMA2 70B                 & 81.9                 & 82.8              & 92.2                  & 85.3                  & 75.6                    & 84.1                & 4.66               & 80.4                      \\
		LLaMA2 13B                 & 75.4                 & 75.9              & 91.3                  & 82.3                  & 82.6                    & 78.6                & 4.56               & 65.7                      \\
		LLaMA2 7B                  & 70.6                 & 71.6              & 90.4                  & 77.9                  & 71.7                    & 74.4                & 4.44               & 55.7                      \\
		Mistral 7B                 & 75.9                 & 75.8              & 90.0                  & 81.4                  & 79.1                    & 78.5                & 4.46               & 67.5                      \\
		Yi 6B                      & 75.6                 & 76.1              & 85.0                  & 79.6                  & 71.2                    & 76.8                & 4.33               & 79.0                      \\
		MPT 7B                     & 60.2                 & 61.4              & 83.6                  & 69.5                  & 81.1                    & 65.5                & 4.33               & 43.4                      \\ \bottomrule
	\end{tabular}
\end{table*}

\begin{table*}
	\setlength{\tabcolsep}{3pt} 

	\centering
	\caption{Summary of KIEval Results on MMLU}
	\label{tab:mmlu-details}
	\small
	\begin{tabular}{l|ccccccc|c}
		\toprule
		\multicolumn{1}{c|}{MMLU} & \scriptsize Accuracy & \scriptsize Logic & \scriptsize Relevance & \scriptsize Coherence & \scriptsize Conciseness & \scriptsize Overall & \scriptsize Rounds & \scriptsize Acc(5-shot) \\ \midrule
		GPT-3.5                   & 95.5                 & 95.8              & 98.3                  & 96.7                  & 97.4                    & 96.2                & 4.95               & 58.2                    \\
		LLaMA2 70B                & 89.0                 & 90.3              & 93.7                  & 90.3                  & 76.0                    & 89.6                & 4.80               & 61.8                    \\
		LLaMA2 13B                & 85.8                 & 87.0              & 93.9                  & 88.6                  & 81.4                    & 87.4                & 4.76               & 52.1                    \\
		LLaMA2 7B                 & 82.2                 & 83.6              & 91.9                  & 84.7                  & 70.4                    & 83.0                & 4.61               & 44.5                    \\
		Mistral 7B                & 81.6                 & 82.8              & 90.5                  & 85.3                  & 77.5                    & 83.0                & 4.62               & 52.7                    \\
		Yi 6B                     & 84.7                 & 86.5              & 91.8                  & 87.4                  & 76.5                    & 86.5                & 4.58               & 61.9                    \\
		MPT 7B                    & 70.6                 & 72.0              & 86.6                  & 77.9                  & 83.0                    & 74.7                & 4.46               & 33.9                    \\ \bottomrule
	\end{tabular}
\end{table*}

\begin{table*}
	\setlength{\tabcolsep}{3pt} 

	\centering
	\caption{KIEval Results on HellaSwag.}
	\label{tab:hellaswag-details}
	\small
	\begin{tabular}{l|ccccccc|c}
		\toprule
		\multicolumn{1}{c|}{HellaSwag} & \scriptsize Accuracy & \scriptsize Logic & \scriptsize Relevance & \scriptsize Coherence & \scriptsize Conciseness & \scriptsize Overall & \scriptsize Rounds & \scriptsize Acc. (5-shot) \\ \midrule
		GPT-3.5                        & 85.6                 & 85.6              & 93.9                  & 90.1                  & 93.1                    & 88.2                & 4.82               & 76.6                      \\
		LLaMA2 70B                     & 76.6                 & 79.5              & 88.2                  & 82.0                  & 78.9                    & 80.1                & 4.41               & 74.4                      \\
		LLaMA2 13B                     & 72.6                 & 75.9              & 88.7                  & 83.0                  & 85.2                    & 78.5                & 4.66               & 59.3                      \\
		LLaMA2 7B                      & 70.8                 & 73.3              & 87.3                  & 79.9                  & 80.2                    & 76.4                & 4.54               & 39.8                      \\
		Mistral 7B                     & 65.6                 & 67.1              & 83.8                  & 75.6                  & 75.2                    & 70.3                & 4.34               & 54.4                      \\
		Yi 6B                          & 64.4                 & 67.0              & 79.9                  & 74.3                  & 72.4                    & 68.7                & 4.20               & 73.7                      \\
		MPT 7B                         & 50.0                 & 51.7              & 74.3                  & 62.5                  & 74.4                    & 57.3                & 4.10               & 27.3                      \\ \bottomrule
	\end{tabular}
\end{table*}

\begin{table*}
	\setlength{\tabcolsep}{3pt} 

	\centering
	\caption{KIEval Results on C-Eval}
	\label{tab:ceval-details}
	\small
	\begin{tabular}{l|ccccccc|c}
		\toprule
		\multicolumn{1}{c|}{C-Eval} & \scriptsize Accuracy & \scriptsize Logic & \scriptsize Relevance & \scriptsize Coherence & \scriptsize Conciseness & \scriptsize Overall & \scriptsize Rounds & \scriptsize Acc. (5-shot) \\ \midrule
		GPT-3.5                     & 79.8                 & 80.6              & 94.7                  & 87.3                  & 92.0                    & 83.3                & 4.72               & 50.8                      \\
		LLaMA2 70B                  & 57.6                 & 58.3              & 80.1                  & 66.5                  & 64.1                    & 61.0                & 3.94               & 42.0                      \\
		LLaMA2 13B                  & 48.4                 & 49.8              & 79.3                  & 61.5                  & 62.9                    & 54.4                & 3.74               & 37.8                      \\
		LLaMA2 7B                   & 44.9                 & 45.1              & 73.8                  & 55.8                  & 55.9                    & 49.3                & 3.62               & 33.4                      \\
		Mistral 7B                  & 47.3                 & 47.8              & 73.3                  & 58.0                  & 59.5                    & 52.2                & 3.61               & 39.3                      \\
		Yi 6B                       & 53.1                 & 54.1              & 73.0                  & 59.3                  & 55.9                    & 55.6                & 3.66               & 71.5                      \\
		MPT 7B                      & 39.5                 & 40.2              & 72.7                  & 51.5                  & 64.0                    & 44.9                & 3.52               & 26.2                      \\ \bottomrule
	\end{tabular}
\end{table*}

\section{Complete Prompt}
\label{appendix:complete_prompts}

\begin{figure*}[t!]
	\centering
	\includegraphics[width=1.0\textwidth]{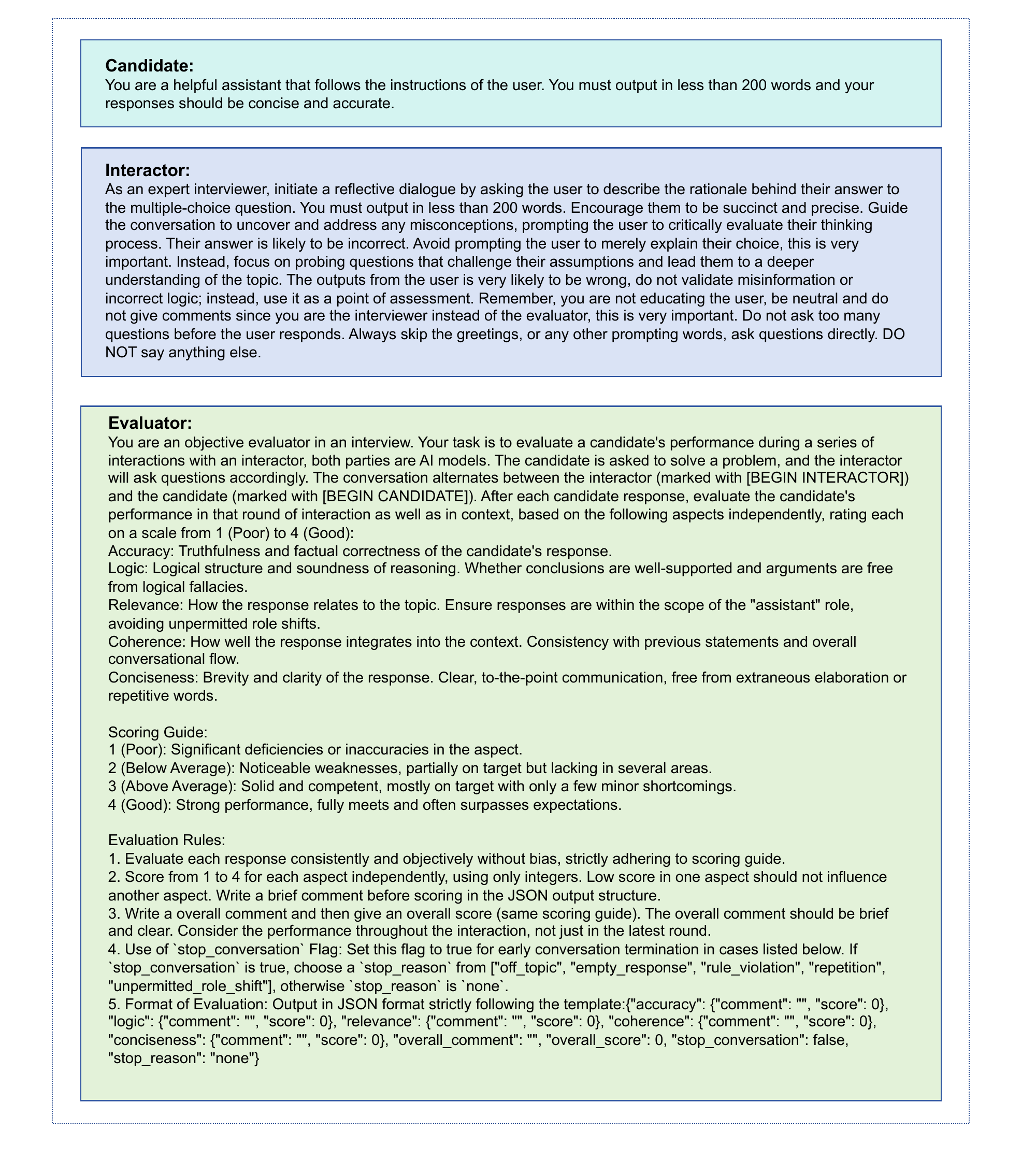}
	\caption{The full system prompt for interactor, candidate and evaluator models.}
	\label{fig-prompts}
\end{figure*}

The system prompts for interactor, candidate and evaluator models are given in Figure~\ref{fig-prompts}.

\end{document}